\documentclass{article}
\usepackage{PRIMEarxiv}
\usepackage[utf8]{inputenc} 
\usepackage[T1]{fontenc}    
\usepackage{hyperref}       
\usepackage{url}            
\usepackage{booktabs}       
\usepackage{amsfonts}       
\usepackage{nicefrac}       
\usepackage{microtype}      
\usepackage{lipsum}
\usepackage{fancyhdr}       
\usepackage{graphicx}       
\graphicspath{{media/}}     

\title{
  Neural Machine Translation For Low Resource Languages \\
}

\author{
  Vakul Goyle \\
  School of Computer Science and Engineering \\
  Nanyang Technological University \\
  \texttt{vakul002@e.ntu.edu.sg} \\
  \And
  Parvathy Krishnaswamy \\
  School of Computer Science and Engineering \\
  Nanyang Technological University \\
  \texttt{parvathy002@e.ntu.edu.sg} \\
  \AND
  Kannan Girija Ravikumar\\
  School of Computer Science and Engineering \\
  Nanyang Technological University \\
  \texttt{kannan006@e.ntu.edu.sg} \\
  \And
  Utsa  Chattopadhyay \\
  School of Computer Science and Engineering \\
  Nanyang Technological University \\
  \texttt{utsa001@e.ntu.edu.sg} \\
  \AND
  Kartikay Goyle \\
  CodeAssist, Inc.  \\
  Chief Technology Officer \\
  \texttt{kartikay@codeassist.ai} \\
}

\begin{document}

\maketitle

\begin{abstract}
Neural Machine translation is a challenging task due to the inherent complex nature and the fluidity that natural languages bring. Nonetheless, in recent years, it has achieved state-of-the-art performance in several language pairs. Although, a lot of traction can be seen in the areas of multilingual neural machine translation (MNMT) in the recent years, there are no comprehensive survey done to identify what approaches work well. The goal of this paper is to investigate the realm of low resource languages and build a Neural Machine Translation model to achieve state-of-the-art results. The paper looks to build upon the \texttt{mBART.CC25} \cite{liu2020multilingual} language model and explore strategies to augment it with various NLP and Deep Learning techniques like back translation and transfer learning. This implementation tries to unpack the architecture of the NMT application and determine the different components which offers us opportunities to amend the said application within the purview of the low resource languages problem space.
\end{abstract}


    

\section{Introduction}

Availability of large parallel corpora for training has helped in significant development and advancement in the field of MT. Modern neural machine translation techniques can achieve near human-level translation performance for those language pairs where sufficient parallel training resources exist (e.g. English $\longrightarrow$ French, English $\longrightarrow$ German, and vice-versa). The neural models often seemed to perform poorly on low resource language pairs because of low volume of parallel data. Additionally, it is also difficult to evaluate models trained on such language pairs, because of the lack of publicly available benchmarks. In short, there are several challenges to solve in order to improve translation for low resource languages; scarcity of clean parallel data, alternative sources like monolingual resources, noisy comparable data, parallel data in related languages, etc. to name a few.

To cater these problems, in this work we have taken the \texttt{mBART.CC25} pretrained language model \cite{liu2020multilingual} and created a baseline model for four languages, namely Sinhala $\longrightarrow$ English, Nepali $\longrightarrow$ English, Khmer $\longrightarrow$ English and Pashto $\longrightarrow$ English, from \textbf{FLoRes} \cite{guzman2019flores} datasets. 

\newpage

Over this generated baseline, we applied techniques like transfer learning, back translation and introducing loss criterion using focal loss, and verified their effect on \textbf{BLEU} \cite{papineni-etal-2002-bleu} score. Extensive experiments on these approaches shows that our simple approach works prodigiously well and further improvements from the back translation (BT) \cite{sennrich-etal-2016-improving} and replacing cross entropy with focal loss \cite{lin2018focal} sets a new state-of-the-art on both \emph{Si-En} and \emph{Ne-En}, improving the results by \textbf{0.57} and \textbf{1.29} respectively over \texttt{mbART.CC25} \cite{liu2020multilingual} fine tuned model for NMT. We also experimented on language pairs \emph{Km-En} and \emph{Ps-En}, recent additions in \textbf{FLoRes} but with no publically available benchmarks; and got new benchmark of \textbf{6.68} and \textbf{1.14} respectively. Finally we present our new NMT model which can perform across the language translation and some interactive translation phenomenon too.

\section{Related Work (or Background)}

\subsection{mBART pretrained}
For this paper, the idea was to piggyback on the different work that is already been done around multilingual NMT and with that in mind, we used the \texttt{mBART.CC25} \cite{liu2020multilingual} model which is trained and fine tuned on several of the low and high resource languages. mBART is a successor of Facebook's BART model that uses bi-directional and auto-regressive models. The mBART model is fine tuned to provide support for multiple languages for machine translation purposes. It is trained on the monolingual \emph{common crawl} corpus for 25 languages (CC25). The text corpus is further tokenized using the sentence-piece model (SPM) \cite{kudo2018sentencepiece} which provides both subword tokenization and unigram language modelling additionally including direct training over complete sentences. mBART further uses up/down sampling to balance out the percentage of each languages. This pretrained model is then further fine tuned on parallel data corpus for 24 languages that achieved state-of-the-art results for many of them. 

The process of collecting corpora for the low resource languages was the next hurdle. The \textbf{FloRes} \cite{guzman2019flores} dataset was the obvious choice and the published work around that was a good starting point. Additionally we used the \textbf{OPUS} datasets \cite{TIEDEMANN12.463} for gathering parallel datasets for use cases which did not have sufficient datasets in FLoRes. For Back translation we used the monolingual dataset from \textbf{SMT dataset} \cite{statmt} in \texttt{http://statmt.org/}.

\subsection{Focal Loss}

Auto-regressive sequence-to-sequence (seq2seq) models such as Transformers \cite{vaswani2017attention} are trained to maximize the log-likelihood of the target sequence, conditioned on the input sequence. However, seq2seq models (or structured prediction models in general) suffer from a discrepancy between token level classification during learning and sequence level inference during the search. This discrepancy also manifests itself in the form of the curse of sentence length, i.e. the models’ proclivity to generate shorter sentences during inference, which has received considerable attention in the literature (\emph{Pouget-Abadie et al., 2014}; \cite{pouget2014overcoming}; \emph{Murray and Chiang,2018} \cite{murray2018correcting}).

For transformer based NMTs, tokens with low frequency receive lower probabilities during prediction because the embeddings for low frequency tokens lie in a different subregion of space than semantically similar high frequency tokens. Furthermore, since these token embeddings have to match the context vector for getting the next token probabilities, the dot-product similarity score is lower for low frequency tokens, even when they are semantically similar to high frequency tokens. 

In this work, we propose using the focal loss function to better incorporate the inductive biases compared to widely used cross entropy (CE) loss. This is because cross entropy loss limits NMT models’ expressivity during inference and isn't able to generate low frequency tokens in the output effectively. Proposed in (\emph{Lin et al., 2017} \cite{lin2017focal}),  Focal loss  (FL)  increases the relative loss of low-confidence predictions vis-à-vis high-confidence predictions when compared to cross entropy.

The focal loss computation is described in equation \ref{eqn:focal_loss}, where $\gamma$ is the non-negative ($> 0$) tuneable focusing parameter, $p$ is the probability/confidence of the prediction and $\alpha$ is the parameter that balances the importance of positive/negative samples to deal with the class imbalance.

\begin{equation}
\mathrm{FL}\left(p_{\mathrm{t}}\right)=-\alpha_{\mathrm{t}}\left(1-p_{\mathrm{t}}\right)^{\gamma} \log \left(p_{\mathrm{t}}\right)
\label{eqn:focal_loss}
\end{equation}

\newpage
\subsection{Back Translation}

Circling back to issues of unavailability of large parallel datasets for training for low resource languages, we come across the approach of a synthetic data generation method called back translation proposed by \emph{Sennrich, Haddow and Birch} \cite{sennrich-etal-2016-improving}. As stated in \emph{Sennrich, Haddow and Birch} \cite{sennrich-etal-2016-improving}, encoder-decoder NMT architectures already can learn the same information as a language model, and we explore strategies to train with monolingual data without changing the neural network architecture. In this method, a translator in the opposite direction, i.e. from target to source, will be trained. Using the trained translator, a large set of target monolingual data is translated to source language resulting in a synthetic parallel corpus of source and target languages. The generated synthetic data will be added to the already existing source to target parallel corpus and is used to train the translator model in the source to the target direction. Combining the existing parallel corpus with the synthesized data gives a large dataset that can be used for training.

\begin{figure}[hbt!]
  \centering
  \includegraphics[scale=0.6]{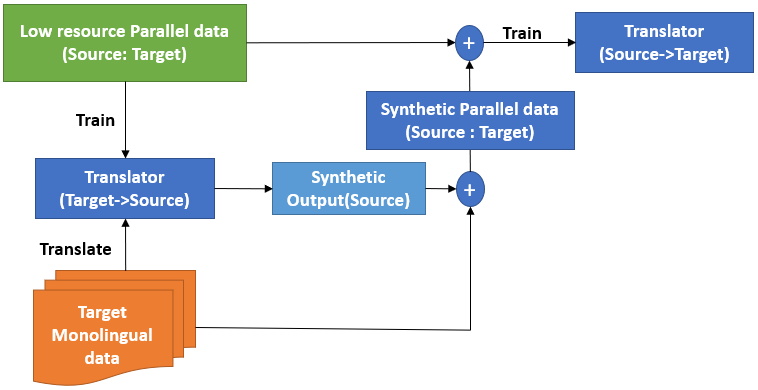}
  \caption{Back Translation flowchart}
  \label{fig:reviews-length}
\end{figure}

A different variant of back translation is suggested in \emph{Hoang et. al.} \cite{hoang-etal-2018-iterative} called iterative back translation. In this method the process of back translation is repeated multiple times, by appending the synthesized data on each iteration, producing a large parallel synthesized dataset.

\subsection{Transfer Learning}
Another approach proposed by many recent works is around Transfer learning. Transfer learning method can significantly improve the BLEU score across a range of low resource languages as mentioned in \emph{Zoph et al.} \cite{zoph-etal-2016-transfer}. The key idea is to first to train a high resource language pair to generate the parent model, then transfer some of the learned parameters to the low resource pair's child model, to initialize and constrain training. NMT systems have achieved competitive accuracy rates under large data training conditions for language pairs such as English $\longrightarrow$ French. However, neural methods are data-hungry and learn poorly from low count events. This behaviour makes vanilla NMT a poor choice for low resource languages, where parallel data is scarce. Therefore, transfer learning uses knowledge from a learned task to improve the performance on a related task, typically reducing the amount of training data required as mentioned in articles \cite{Torrey2009Chapter1T} and \cite{5288526}.

As a matter of fact, we have also taken this approach in our work and used it. We first train an NMT model on a large corpus of parallel data (e.g., French $\longrightarrow$ English). We call this the parent model. Next, we initialize an NMT model with the already trained parent model. 

This new model is then trained on a very small parallel corpus (e.g., Uzbek $\longrightarrow$ English). We call this the child model. Rather than starting from a random position, the child model is initialized with the weights from the parent model. A justification for this approach is that we need a strong prior distribution over models in scenarios where we have limited training data. The parent model trained on a large amount of bilingual data can be considered an anchor point, the peak of our prior distribution in model space. 

When we train the child model initialized with the parent model, we fix parameters likely to be useful across tasks so that they will not be changed during child model training. In the French $\longrightarrow$ English to Uzbek $\longrightarrow$ English example, as a result of the initialization, the English word embeddings from the parent model are copied, but the Uzbek words are initially mapped to random French embedding as cited in \emph{Zoph et. al.} \cite{zoph-etal-2016-transfer}.

There has also been work on using data from multiple language pairs in NMT to improve performance. Actually, in \cite{dong-etal-2015-multi} the authors showed that using this framework improves performance for low resource languages by incorporating a mix of low resource and high resource languages. 

 \section{Approach (or Method)}

\subsection{Baseline Model}
Given the time and resource constraint and the substantial benefits, we decided to use the pretrained model. This decision was only furthered by the nature of the languages we chose for this assignment. All the languages in question followed a similar morphology, syntax and linguistic topology and belonged to either the Indo-Aryan language family or a close relative. They also followed a similar subject-object-verb (\emph{SOV}) structure with the exception of Khmer which followed a subject-verb-object (\emph{SVO}) linguistic topology. Barring this slight difference, the nature of our task still was quite similar to what can be accomplished using the \texttt{mBART-CC25} pretrained model. Besides, the 25 languages chosen for training \textbf{mBART} included \emph{Sinhala(si)} and \emph{Nepali(ne}, which made \textbf{mBART} a good starting point for them. Additionally, \textbf{mBART} was also trained in \emph{Vietnamese(vi} which is closer to \emph{Khmer(km} which share a similarity score of 11\%, besides both being isolating languages with no inflectional morphology.

For generating the baseline for this experiment, we used \texttt{mBART.CC25} and fine tuned it for the task of translation on all the 4 languages separately using the \textbf{FLoRes} and \textbf{OPUS} parallel corpus and generated 4 different baselines. We used \texttt{label\_smoothed\_cross\_entropy} loss criterion with a label smoothing threshold of 0.2. We also relied on the \texttt{sentence-piece} model and the dictionary provided by \texttt{mBART.CC25} itself. 

The obtained model is set as the baseline model of that particular language. The experimentation of different techniques were done over the baseline model. We checked whether the experimented techniques are able to improve the model performance compared to the baseline model. If the performance is improved, then we conclude that it can be used as a performance enhancing technique for low resource languages.

\subsection{Focal Loss}
We proposed Focal loss, which would better adapt model training to the structural dependencies of conditional text generation by incorporating the inductive biases in the training process.

For experiments, we used \texttt{fairseq} \cite{ott2019fairseq} library, which did not have support for focal loss, so we wrote our own custom implementation of focal loss based on the implementation proposed by \emph{Raunak et al.,2020} \cite{raunak2020long}, setting the hyper-parameter $\alpha$ to 0.5 and $\gamma$ to 1 respectively. We used the baseline model and fine tuned it further using focal loss on the low resource languages - \emph{Sinhala(si)}, \emph{Nepali(ne)}, \emph{Pashto(ps)} and \emph{Khmer(km)}; to build the final model.

\subsection{Back Translation}
The English dataset created by news crawling, provided by \url{statmt.org} \cite{statmt} was used as the target monolingual dataset for the purpose of Back translation. 

As part of back translation, we fine tuned the \texttt{mBART.CC25} over the parallel corpus in the reverse order, i.e. from the target language to source language. Using the reverse translator, the large target monolingual data was translated to get the respective source sentences as output to create the synthetic parallel corpus data. The quality of this synthesized data depends on the reverse translator's quality. An average of 8 million parallel records was generated for each language.

The synthesized parallel corpus was combined with the existing low resource parallel corpus to create an extensive source-target dataset. The best model was obtained after training with \texttt{focal\_loss} as the loss criterion instead of \texttt{label\_smoothed\_cross\_entropy}, over the synthetic parallel dataset. This somehow solved the problem of having less parallel data since the monolingual dataset is widely available for any languages; thus providing us with a good sizeable corpus to train the model.

\subsection{Transfer Learning}

For experimenting with the knowledge transfer / transfer learning approach, we used the \texttt{mBART.CC25} model and trained them on the relatively high resource languages - \emph{Tamil(ta)}, \emph{Hindi(hi)}, \emph{Vietnamese(vi)} and \emph{Farsi(fa)} and built the parent model. Then, using the parent model we fine tuned it further on the low resource languages in question - \emph{Sinhala(si)}, \emph{Nepali(ne)}, \emph{Pashto(ps)} and \emph{Khmer(km)}; to build the final model. We also used the \texttt{focal\_loss} as the loss criterion instead of \texttt{label\_smoothed\_cross\_entropy}. This affects how the \emph{best} checkpoints (i.e. resultant output model) are identified.

\section{Experiments}

This section talks about the experiment settings and steps done to implement, understand and evaluate the proposals listed above.

\subsection{Data}
\subsubsection{Datasets}

We trained the \texttt{mBART.CC25} pretrained model on the 4 languages available in the \textbf{FLoRes} repository. The training data parallel corpus for \emph{Ps-En} and \emph{Km-En} are sourced from the \textbf{OPUS} data library. The monolingual corpus of the languages are sourced from \url{statmt.org}'s \cite{statmt} new article corpus. While, the parallel data for the parent languages used for transfer learning is also sourced from the \textbf{OPUS} library. We primarily relied on the \texttt{GNOME} \cite{gnome}, \texttt{KDE4} \cite{kde4}, \texttt{Ubuntu} \cite{ubuntu} localization corpus and \texttt{OpenSubtitles} \cite{opensub} movie subtitles corpus / \texttt{Global Voices} \cite{casmacat} news articles corpus for training, which gives a good mix of short as well as long sentences. The corpus used for validation and test was \texttt{Wikipedia} \cite{WOLK2014126}.

The details of the corpus size can be referred in Table \ref{table:datasets}

\begin{table}[hbt!]
\centering
\caption{Datasets}
\label{table:datasets}
\begin{tabular}{l|rrr|rrr}
\hline
 &
  \multicolumn{1}{l}{\textit{Train}} &
  \multicolumn{1}{l}{\textit{Valid}} &
  \multicolumn{1}{l|}{\textit{Test}} &
  \multicolumn{1}{l}{\textit{Train}} &
  \multicolumn{1}{l}{\textit{Valid}} &
  \multicolumn{1}{l}{\textit{Test}} \\ \hline
\multicolumn{1}{c|}{\textbf{Languages}}                                                       & \multicolumn{3}{c|}{\textbf{Si-En}} & \multicolumn{3}{c}{\textbf{Ne-En}} \\ \hline
\textbf{\begin{tabular}[c]{@{}l@{}}Baseline training \&\\ Fine tuning\end{tabular}}           & 401490      & 2898      & 2766      & 563025     & 2559       & 2835     \\ \hline
\textbf{Back translation}                                                                     & 1157650      &  2898     &   2766     &     1484945      &    2559        &    2835      \\ \hline
\textbf{\begin{tabular}[c]{@{}l@{}}Transfer learning\\ (parent parallel corpus)\end{tabular}} & 143638      & 93540     & 11507     & 536137     & 22129      & 65100    \\ \hline
\multicolumn{1}{c|}{\textbf{Languages}}                                                       & \multicolumn{3}{c|}{\textbf{Ps-En}} & \multicolumn{3}{c}{\textbf{Km-En}} \\ \hline
\textbf{\begin{tabular}[c]{@{}l@{}}Baseline training \&\\ Fine tuning\end{tabular}}           & 108333      & 3162      & 2698      & 127727     & 2378       & 2309     \\ \hline
\textbf{Back translation}                                                                     &   931053          &    3162       &    2698       &     871887       &   2378   & 2309 \\ \hline
\textbf{\begin{tabular}[c]{@{}l@{}}Transfer learning\\ (parent parallel corpus)\end{tabular}} & 7341252     & 628192    & 225878    & 3553218    & 12732403   & 381408   \\ \hline
\end{tabular}
\end{table}
 
\subsubsection{Preprocessing}
Similar to \texttt{mBART.CC25} we use the \texttt{sentence-piece model (SPM)} for tokenization. The datasets were pre-processed to exclude any sentences that were too small - i.e. one token long sentences for \emph{Nepali(ne)}, \emph{Pashto(ps)} and \emph{Khmer(km)} and sentences with fewer than 6 tokens for \emph{Sinhala(si)}. We used the \texttt{fairseq} library \cite{ott2019fairseq} to run the preprocessing as well as training, but instead of relying on the \texttt{sentence-piece} tokenization model provided by the \textbf{FLoRes} repository, we used the one provided by the \texttt{mBART.CC25}.

\subsection{Evaluation method}

We have used the sacreBLEU metric for the evaluation method to get the mentioned results. Comparing BLEU scores is harder than it should be, as mentioned in \emph{Papineni et al., 2002} \cite{papineni-etal-2002-bleu}. Every decoder has its own implementation, often borrowed from moses \cite{moses}, but maybe with subtle changes. Moses itself has a number of implementations as standalone scripts, with little indication of how they differ. Different flags passed to each of these scripts can produce wide swings in the final score. All of these may handle tokenization in different ways. So what sacreBLEU does is that it decodes and detokenizes the sentences before passing it to \textit{corpus-bleu}. SacreBLEU aims to solve these problems by wrapping the original (\emph{Papineni et al., 2002} \cite{papineni-etal-2002-bleu}) reference implementation together with other useful features. 

The defaults are set the way that BLEU should be computed, and the script outputs a short version string that allows others to know precisely what has been done. NLTK and SacreBLEU use different tokenization rules, mainly in handling punctuation. NLTK uses its tokenization, whereas SacreBLEU replicates the original Perl implementation \cite{papineni-etal-2002-bleu}. The tokenization rules are probably more elaborate in NLTK, but they make the number incomparable with the original implementation.

For our evaluation, we used the \texttt{fairseq-generate} CLI with SacreBLEU as the scoring method. Additionally we also used the \texttt{lenpen} hyper-parameter suggested in \emph{Yinhan et al., 2002} \cite{liu2020multilingual} with a value of $1 2$, so that it favors the longer sentences. This approach gave us a better BLEU score.

\subsection{Model settings}

We opted for the \texttt{mbart\_large} architecture which follows the same settings of the \texttt{bart-large} - i.e. 24-layer, 1024-hidden, 16-heads, but intead of 400M parameters, \texttt{mbart\_large} has 610M parameters (num. trained: 610851840). We chose a learning rate of 0.0003, a dropout ratio of 0.3 and ADAM \cite{kingma2017adam} optimizer for training. Since the corpus size was very small, we opted to use the \texttt{max\_updates} argument of the \texttt{fairseq-train} command to determine the stopping criteria for training. \texttt{updates} expresses the number of times the parameter values are learned and tuned. We chose to stop at 40,000 \texttt{updates}. Resource wise, we chose 12 NVIDIA's Tesla T4 GPU machines to train for 1 week for all 4 languages.

\subsection{Results}

\begin{table}[hbt!]
\begin{center}
\caption{BLEU Score Results}
		\label{table:result}
		\begin{tabular}{l|c c c c}
			\hline			 
			\textbf{-} & \textbf{SI-EN} &\textbf{NE-EN} &\textbf{KM-EN} &\textbf{PS-EN}\\
			\hline 
			State-of-the-art (mBART) & 13.7 & 14.5 & - & - \\
			
		    Baseline Model & 12.3 & 11.53 & 1.45 & 0.46  \\
		   
		    Focal Loss & 12.90 & 12.97 & 5.53 & 0.76  \\
		    
		    Back Translation with Focal Loss & \textbf{14.27} & \textbf{15.79} & \textbf{6.18} & \textbf{1.14} \\
		    
		    Transfer Learning with Focal Loss & 12.20 & \textbf{14.62} & \textbf{6.49} & \textbf{2.07} \\
		    \hline 
 		\end{tabular}	
	\end{center}
\end{table}     

From the Table \ref{table:result} it can be observed that the baseline model did not outperform the state-of-the-art model (\texttt{mBART}), mainly because we trained our baseline model primarily with GNOME, KDE4 and Ubuntu localization files corpus which consist of short sentences (i.e. sentences with fewer tokens), but the validation/testing was done on the Wikipedia corpus comprising of longer sentences. Adding to this, the Wikipedia corpus also caused a domain drift, resulting in a low BLEU score as compared to state-of-the-art.

The experiment on focal loss showed the improvement over the baseline model but could not beat the state-of-the-art model. Although it fulfilled our expectations of outperforming the baseline model and confirmed our hypothesis that focal loss is a better loss over cross-entropy for low resource languages, the results were still not better than the state-of-the-art model. This could be attributed to the fact that the fine-tuning for focal loss was done on the baseline model, whose performance was already impacted due to the quality and domain drift observed in the corpus.

With back translation we observed that the model outperformed both the baseline and the state-of-the-art as highlighted in bold in Table \ref{table:result}. This was mainly because in the case of back translation the synthetic parallel data generated from the monolingual corpus was at least five times larger than existing parallel data which was appended to the final training data, resulting in improved model performance. 

Furthermore, in the case of transfer learning, we observe that language pairs - \emph{Ne-En}, \emph{Km-En}, \emph{Ps-En} outperformed the baseline as well as a state-of-the-art model as highlighted in bold in Table \ref{table:result}. But \emph{Si-En} turned out to be an exception to the rule. This outcome can be attributed to the scarcity of clean, annotated corpus for Tamil. The \emph{Ta-En} parallel corpus was relatively smaller and also contained many small (fewer tokens) sentences, resulting in not augmenting the performance in \emph{Si-En}.

The three languages' (\emph{Ne-En}, \emph{Km-En}, \emph{Ps-En}) BLEU score results indeed indicate that our hypothesis, approach and methodologies are correct at handling the problems of translation of low resource languages, but it also highlighted the significance of data quality which can be observed in \emph{Si-En} - Transfer Learning with Focal Loss experimentation, where the quality of data had a bigger influence on the outcome than the methodology.

\newpage
\section{Analysis}
\begin{figure}[ht!]
  \centering
  \includegraphics[scale=0.25]{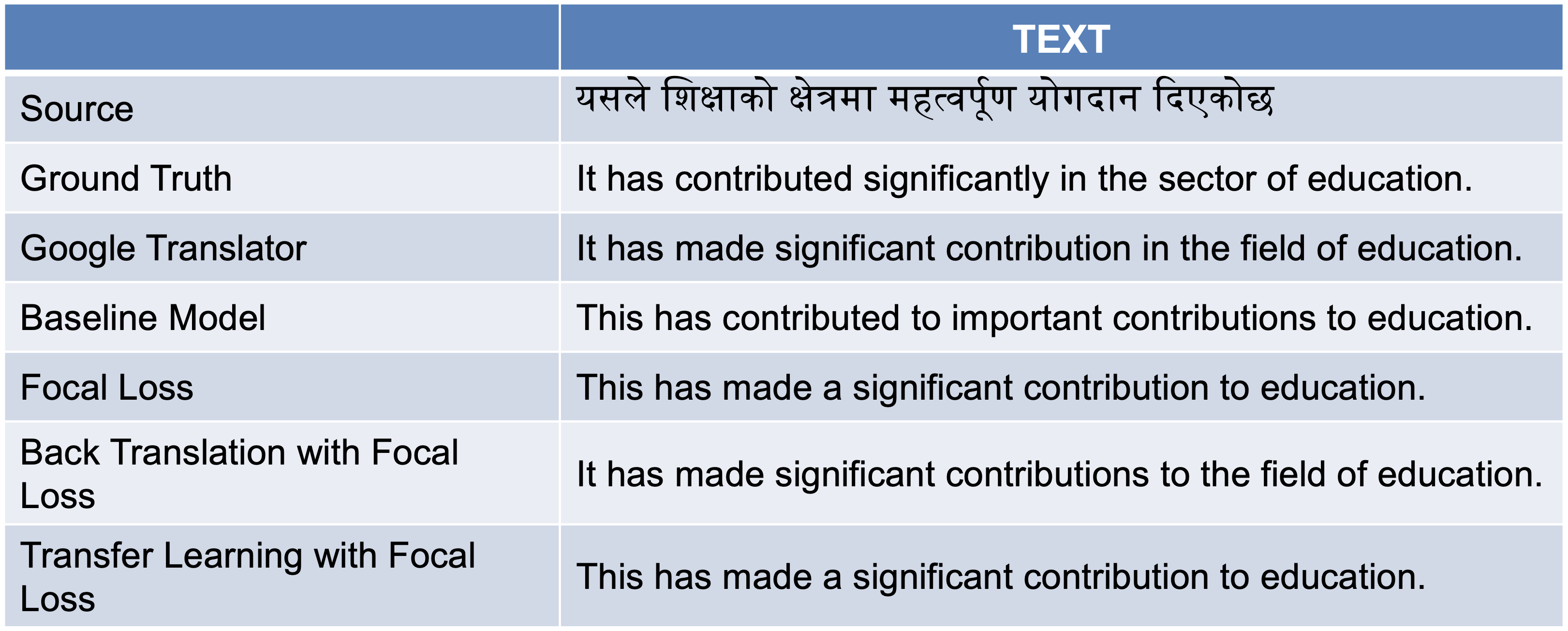}
  \caption{Comparison of translation.}
  \label{fig:tln_comp}
\end{figure}

The trained models' translation output is analyzed using \texttt{fairseq-generate} and \texttt{fairseq-interactive}, the two command-line tools provided by the \texttt{fairseq} package. \texttt{fairseq-interactive} converts the raw source text one at a time to target using the trained model while \texttt{fairseq-generate} translates a batch or a file of pre-processed source data to target data using the trained model. After training, translation is performed with the help of these tools. Finally, the resulting translation is compared with the output of google translate to understand how good is the model translation. An example of Nepali to English translation comparison can be seen in  Figure \ref{fig:tln_comp}.

From the comparison in Figure \ref{fig:tln_comp}, it can be observed that the baseline model translation is the least accurate one. The translation of the back translation model is similar to the google translator and ground truth. Comparing BLEU scores of the four trained languages, it can be observed that the languages \emph{Nepali(ne)} and \emph{Sinhala(si)} has a significantly better score compared to transfer learning. 

One of the reasons for this can be that the model \texttt{mBART.CC25} is pretrained using the languages \emph{Nepali(ne)} and \emph{Sinhala(si)} and not with the other two. The model will better understand the languages used for pre-training, thus helping to perform better in those languages.

\section{Conclusion}
In this paper, we demonstrated the benefits of leveraging the pretrained models. We observed that the training iterations required for building the baseline over a pretrained model was significantly lower as compared to \texttt{mBART.CC25} which was trained over 10x more iterations. Although, this still ended up being resource intensive and required substantial training time. Additionally, we were able to demonstrate the benefits of applying back translation as well as transfer learning with focal loss to overcome the limitations of low resource languages; which can be observed in the scores. The scores observed for these models surpassed the state-of-the-art models for \emph{Nepali(ne)} and \emph{Sinhala(si)}. For \emph{Khmer(KM)} and \emph{Pashto(PS)}, there are no baselines published, and hence we are the first ones to create the state-of-the-art benchmark, which can be used as a baseline to further improvements in future research. 

\subsection{Future work}
All the experiments performed during this paper were done in silos. It would be interesting to look at what improvements can be seen by augmenting them over each other, meaning that applying back translation over the model trained via transfer learning could improve the score. Alternatively, another avenue to look at is the iterative approach cited in \emph{Hoang et. al.} \cite{hoang-etal-2018-iterative}, where we can perform multiple iterations of back translations over multiple monolingual corpora, which significantly increases the training data size. A similar thought process can be extended towards transfer learning as well. Instead of training over a single high resource language, we can explore the possibility of training over a series of high resource languages that share similarities with the language in question, yielding better results as proposed in the paper by \emph{Dong, Wu, He, Yu \& Wang} \cite{dong-etal-2015-multi}.

As an ending note, we would like to draw attention towards the fact that the recent work around MNMT, although has significantly increased, but there still exist a lot of factors that can be deterrent to its progress; for example lack of human evaluation. Looking at avenues to get more people involved will help accelerate the work done on MNMT.


\bibliographystyle{unsrt}
\bibliography{ref}

\appendix


\end{document}